\documentclass[sigconf]{acmart}

\usepackage{amsmath}
\usepackage{booktabs}
\usepackage{enumitem}
\usepackage{graphicx}
\usepackage{multirow}
\usepackage{siunitx}
\usepackage{subcaption}
\usepackage{tikz}
\usepackage{pgfplots}
\usetikzlibrary{arrows.meta,positioning,shapes.geometric,calc,fit}
\pgfplotsset{compat=1.18}
\definecolor{deepseek}{HTML}{4E79A7}
\definecolor{gpt}{HTML}{F28E2B}
\definecolor{gemini}{HTML}{59A14F}
\definecolor{failure}{HTML}{E15759}
\definecolor{neutral}{HTML}{9AA0A6}

\AtBeginDocument{}

\makeatletter
\def\@typeset@author@bx{\bgroup\hsize=\author@bx@wd
  \global\setbox\author@bx=\vtop{\centering
    \@authorfont
    \begingroup
      \let\savedpar\par
      \@tempcnta=\z@
      \def\par{\advance\@tempcnta\@ne
        \ifnum\@tempcnta=\@ne\relax
        \else\ifnum\@tempcnta=\tw@,\space
        \else\savedpar\fi\fi}%
      \@currentauthors
    \endgroup
    \par
    \@affiliationfont
    \begingroup
      \let\savedpar\par
      \@tempcnta=\z@
      \def\par{\advance\@tempcnta\@ne
        \ifnum\@tempcnta=\@ne,\space
        \else\savedpar\fi}%
      \@currentaffiliation
    \endgroup}
  \egroup
  \box\author@bx\hspace{\author@bx@sep}%
  \gdef\@currentauthors{}%
  \gdef\@currentaffiliation{}}
\makeatother

\setcopyright{none}
\renewcommand\footnotetextcopyrightpermission[1]{}
\acmYear{2026}
\copyrightyear{2026}

\begin{document}

\title{The Price of Thought: Does Test-Time Reasoning Pay in LLM Trading?}

\author{Jiayi Chen}
\email{jc2693@njit.edu}
\author{Guiling Wang}
\email{guiling.wang@njit.edu}
\affiliation{%
  \department{Department of Computer Science}
  \institution{New Jersey Institute of Technology}
  \city{Newark}
  \state{New Jersey}
  \country{USA}}

\begin{abstract}
While inference-time reasoning in large language models (LLMs) promises better decision making, its higher computational cost may not yield better economic outcomes. Yet reasoning controls are rarely evaluated as economic interventions, where changes in model outputs must translate into better portfolios after trading costs. We conduct a controlled study of representative LLMs from the DeepSeek, GPT, and Gemini families. We vary reasoning effort while holding information available at each formation date, prompts, output formats, and portfolio construction fixed. Our evaluation covers a full year of U.S. equities under three input conditions: numerical, identifiable news, and masked news. It includes more than 800,000 asset predictions and repeated model generations. Across all three model families, additional reasoning does not produce a reliable improvement in net portfolio returns. For DeepSeek, where we examine the full progression from no reasoning to maximum reasoning, performance is nonmonotonic. Repeated generations also produce unstable treatment effects and portfolio selections, even when overall scores remain similar. These findings show that additional reasoning can change financial decisions without reliably improving their economic value, motivating validation for each task before deployment.
\end{abstract}


\maketitle

\makeatletter
\fancyhead[LE]{}
\fancyhead[RO]{}
\makeatother

\section{Introduction}

More reasoning is not necessarily better judgment. A large language model (LLM) can produce a longer, more elaborate analysis and still make the same trade or a worse one. This tension arises with \emph{inference-time reasoning}: extra internal computation requested after a model has been trained, without changing its weights or task input. On benchmarks with verifiable answers, allocating more \emph{test-time compute}, the token budget spent on this extra processing, can improve accuracy by extending a reasoning trajectory, sampling more candidates, or invoking a \emph{verifier} that checks candidate answers \cite{wei2022chain,wang2022selfconsistency,snell2024scaling,muennighoff2025s1}. Trading provides a harder test. Correctness is revealed only later, through noisy prices and after costs, so the relevant question is not whether a response appears more thoughtful, but whether paying for more reasoning improves the resulting decision.

Between the request for more reasoning and the realized investment outcome lies a sequence of transformations:
\begin{center}
\small\textit{Reasoning effort} $\rightarrow$ \textit{stock scores} $\rightarrow$ \textit{holdings} $\rightarrow$ \textit{net returns}
\end{center}
A \emph{stock score} is the model's numerical assessment of a stock's expected return relative to the other candidates. Each link in the sequence can weaken or reverse the apparent benefit of extra computation. A large score change may leave the selected portfolio unchanged, while a small change near a selection boundary may replace a \emph{long} position, which profits when its price rises, or a \emph{short} position, which profits when its price falls. Any gross improvement can then disappear after transaction costs. The effect may also fail to repeat because each model call produces a \emph{stochastic generation}, meaning one sampled output from an otherwise identical request. More reasoning can therefore change scores and holdings without creating stable economic value.

Existing financial LLM studies show that language models can extract information from news, answer financial questions, and participate in trading systems that use tools or multiple agents \cite{lopezlira2023chatgpt,chen2022expected,yu2023finmem,zhang2024finagent,yu2024fincon,xiao2024tradingagents}. These studies establish that systems built around LLMs can trade, but they often change models, prompts, tools, memories, and decision procedures together. They therefore do not isolate what is purchased by increasing reasoning inside an otherwise fixed trader. Research on test-time compute offers limited guidance because it focuses on mathematics, coding, and other tasks with immediate correctness signals \cite{lightman2023verify,snell2024scaling,deepseek2025r1}. Together, these literatures leave a narrower economic question unanswered: \emph{when the model, information, prompt, required output, and trading mechanism are held fixed, does buying more reasoning improve portfolio outcomes?}

\begin{figure*}[t]
  \centering
  \includegraphics[width=\textwidth]{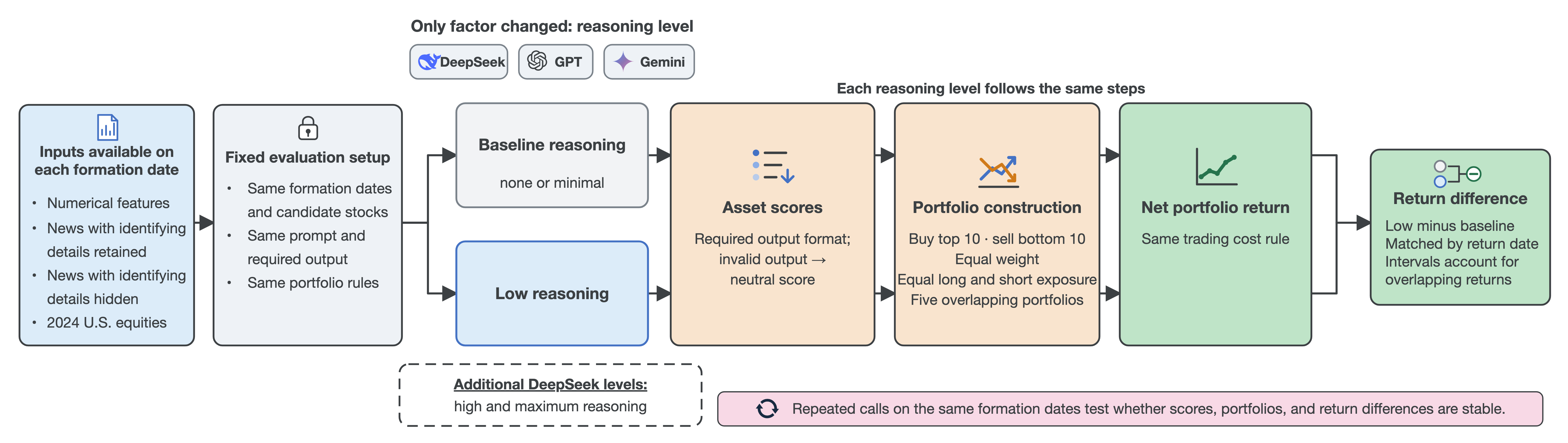}
  \caption{Controlled evaluation pipeline. Within each backbone, reasoning level varies while information available by each formation date, prompts, output schema (required fields, types, and ranges), candidate assets, and portfolio rules remain fixed. The information conditions are numerical inputs, identifiable news with identifying details retained, and masked news with identifying details removed. Repeated calls on frozen formation dates assess the stability of scores, portfolios, and treatment effects.}
  \label{fig:overview}
  \Description{An experimental pipeline sends numerical, identifiable news, or masked news inputs through fixed tasks, two reasoning levels, structured asset scoring, portfolio construction with long and short positions, returns after costs, and a paired estimator of treatment effects.}
\end{figure*}

We answer this question with a controlled intervention on reasoning effort. Within each model family, the reasoning level changes while the information available on each decision date, the prompt, the required output, the candidate stocks, and the portfolio rules remain fixed. Figure~\ref{fig:overview} shows this design as two reasoning paths entering the same scoring, portfolio, cost, and paired evaluation pipeline. Our result is cautionary: added reasoning changes scores, selected assets, failure patterns, and operating cost, but it does not produce a reproducible improvement in return after costs in this setting.

The study follows three predetermined model families over 241 \emph{formation dates} in 2024, where a formation date is a trading day on which new holdings are selected. The common \emph{low versus baseline comparison} raises reasoning from the baseline level to low. The baseline is none for DeepSeek and GPT and minimal for Gemini, whose interface cannot guarantee that reasoning is fully disabled. DeepSeek additionally provides a curve across none, low, high, and maximum reasoning. Each model ranks the same 100 liquid U.S. equities under three information conditions: \emph{numerical} features with identifiers removed; \emph{identifiable news}, which retains company identity; and \emph{masked news}, in which company names and other identifying details are removed. The scores enter the same equal weight, \emph{market neutral} portfolio, which buys the 10 highest ranked stocks, shorts the 10 lowest ranked stocks, and balances the two sides to remove broad market direction. Each selection remains active for five days, creating overlapping \emph{formation cohorts}, or batches of positions opened on different dates. Trading incurs 10 basis points (bps) whenever capital is bought or sold; one bp is 0.01 percentage points.

This controlled design also determines how the evidence must be analyzed. Although the experiment contains 808,800 asset predictions, the predictions and model tasks within a date are not independent economic observations. Primary inference instead uses 241 paired daily portfolio return differences indexed by \emph{return date}, a trading day on which active portfolios earn returns. We apply a Newey--West correction with five lags, a heteroskedasticity and autocorrelation consistent (HAC) method that adjusts uncertainty for changing return variance and serial dependence \cite{neweywest1987}. A predeclared \emph{economic threshold} of $0.25$ bps/day defines the smallest gain considered worthwhile. Invalid tasks receive a neutral score of 0 but remain in the comparison under an \emph{intention-to-treat} rule. To test whether an apparent effect survives another model call, we also conduct \emph{frozen audits}: on 48 fixed formation dates, each backbone receives the same numerical or news task three times under unchanged rules.

The results reveal a consistent difference between behavioral change and economic improvement. None of the nine annual low versus baseline tests establishes a material return benefit, although the generally wide intervals do not show that every effect equals zero. Repeated generations expose a sharper deployment risk. In DeepSeek's three generation audit with masked news, low reasoning reduces return by $9.016$ bps/day relative to none (95\% confidence interval [CI], $-16.248$ to $-1.784$). DeepSeek's response across four reasoning levels is also nonmonotonic, so a favorable adjacent step does not imply that maximum reasoning outperforms no reasoning. GPT and Gemini each change sign across three generations in at least one news condition. Reasoning further changes \emph{output structure reliability}, whether a response follows the required fields and format, in different directions across models. Improved reliability therefore does not by itself imply improved return.

The practical conclusion is asymmetric. Added reasoning has a certain token, monetary, and latency cost, while its economic benefit remains unresolved in nine annual tests and is harmful in one replicated audit. Baseline reasoning is therefore the conservative default unless the target task shows reproducible incremental value. This conclusion is limited to the tested models, inputs, historical year, and portfolio mechanism. It does not imply that reasoning never helps financial decisions, that the estimated effects are statistically equivalent to zero, or that the evaluated baseline portfolios are prospectively profitable. Within that scope, this paper makes three contributions.
\nopagebreak[4]
\begin{enumerate}[topsep=3pt,partopsep=0pt,itemsep=2pt,parsep=0pt]
  \item \textbf{A controlled economic test of reasoning.} We isolate reasoning effort within each backbone while holding the surrounding financial workflow fixed, estimating the marginal value of inference-time compute rather than differences among complete agent systems.
  \item \textbf{Evidence across the decision chain.} We connect structured scores to selected portfolios and returns after costs, then repeat generations to identify reasoning effects that appear favorable once but do not reproduce.
  \item \textbf{Accounting for deployment.} We jointly report return effects, portfolio stability, invalid outputs, token use, and inference cost, revealing when behavioral or reliability changes fail to create economic value.
\end{enumerate}

\section{Related Work}

\subsection{Financial Language Models and Return Prediction}

Financial natural language processing (NLP) progressed from \emph{encoders} adapted to finance, which turn financial text into numerical representations for sentiment and communications \cite{araci2019finbert,yang2020finbert}, to generative models and instruction tuning frameworks for finance \cite{wu2023bloomberggpt,yang2023fingpt}. Evaluation resources have expanded accordingly. FinQA targets numerical reasoning over reports \cite{chen2021finqa}; PIXIU and FinBen cover broad financial understanding and decision tasks \cite{xie2023pixiu,xie2024finben}; and FinTradeBench combines fundamental information with trading signals \cite{agrawal2026fintrade}. These works establish that finance requires both domain knowledge and quantitative reasoning, but benchmark accuracy does not itself show that additional inference-time reasoning improves a portfolio.

A second line of research asks whether representations learned by language models predict returns. Lopez-Lira and Tang find that ChatGPT labels of news headlines forecast subsequent stock movements \cite{lopezlira2023chatgpt}. Chen, Kelly, and Xiu use embeddings from LLMs to form signals that compare expected returns across stocks on the same date \cite{chen2022expected}. These studies motivate our numerical and news conditions, but they also sharpen the distinction that guides our experiment. Representation quality concerns the information encoded by a model; our target quantity, or \emph{estimand}, is the incremental value of spending more compute while holding the representation, prompt, and data fixed.

\subsection{LLM Trading Agents and Financial Benchmarks}

Recent systems embed LLMs in richer trading loops. FinMem introduces layered memory and character design \cite{yu2023finmem}; FinAgent combines multimodal inputs and tools \cite{zhang2024finagent}; FinCon uses a hierarchy of managers and analysts with conceptual verbal reinforcement \cite{yu2024fincon}; and FinRobot provides a general platform for financial agents \cite{yang2024finrobot}. TradingAgents organizes specialized agents into a simulated trading firm \cite{xiao2024tradingagents}. These systems demonstrate breadth, but their components interact: changing reflection, memory, tools, or agent roles changes more than the reasoning level.

Benchmark work increasingly focuses on decision realism. INVESTORBENCH spans financial products and agent tasks \cite{li2024investorbench}. StockBench evaluates sequential decisions while checking \emph{contamination}, meaning possible exposure to evaluation information during training \cite{chen2025stockbench}. From Knowing to Doing masks identifiers to separate memorized market history from decision competence \cite{zhu2026knowing}, while Profit Mirage tests \emph{leakage} (future or evaluation information entering the task) and performance after the model's training data cutoff \cite{li2025profit}. Agentic Trading synthesizes audit concerns for financial agents \cite{xia2026agentic}. We adopt opaque identifiers and masked news for the same reason, while acknowledging that masking cannot prove that historical events were absent from pretraining.

Several recent studies move closer to our question by testing reasoning in financial decisions. Sodha compares direct and thinking LLMs in cross-sectional stock ranking and finds no net portfolio advantage for the thinking model after costs \cite{sodha2025reasoning}. Cheng et al. vary reasoning intensity and market speed in a simulator for tender selection and real time execution, exposing a tradeoff between decision quality and latency \cite{cheng2026agents}. Sitjar toggles a chain-of-thought scaffold within the decision process for a trader of the S\&P 500 exchange-traded fund (SPY) and finds no main improvement in trading performance or \emph{calibration}, the agreement between stated confidence and observed outcomes \cite{sitjar2026reasoning}. These studies show why a controlled reasoning comparison is needed, so we do not claim to be the first to make one. Our contribution is a within backbone design across three model families, common numerical, identifiable news, and masked news tasks, a four level reasoning curve for one backbone, explicit transaction costs, and repeated generation evidence at the score, portfolio, and return levels.

\subsection{Test-Time Compute and Adaptive Reasoning}

\emph{Chain-of-thought prompting} asks a model to produce intermediate reasoning steps and can improve tasks with multiple steps \cite{wei2022chain}. \emph{Self-consistency} samples several such paths and aggregates their answers \cite{wang2022selfconsistency}, while Tree of Thoughts explicitly searches among intermediate possibilities \cite{yao2023tree}. Process supervision and stepwise verification score intermediate steps to select better trajectories \cite{lightman2023verify}. More recent work studies scaling directly at inference. Snell et al. show that test-time compute can substitute for parameter scale when allocation and verification are effective \cite{snell2024scaling}; s1 demonstrates simple budget forcing \cite{muennighoff2025s1}; and DeepSeek-R1 exemplifies models trained for extended reasoning \cite{deepseek2025r1}.

The gains are not free or necessarily monotonic. Think or Not? documents diminishing information gain in long traces and proposes \emph{adaptive stopping}, ending reasoning when extra steps add little information \cite{yong2025think}. Adaptive Model and Strategy Routing lets a system choose a model and reasoning strategy under a budget \cite{pan2025route}, and a recent survey organizes controllable and adaptive methods for test-time compute \cite{alomrani2025budget}. These approaches usually optimize accuracy against a verifier. We instead treat realized portfolio return as the downstream outcome and test the more basic premise that a provider's reasoning control has incremental economic value in a noisy decision pipeline with delayed feedback.

\subsection{Evaluation Under Financial Noise}

Financial evidence requires stronger controls than a static accuracy benchmark. News can move markets \cite{tetlock2007giving}, and machine learning can extract nonlinear return signals \cite{gu2020empirical}, but those relationships are \emph{nonstationary}: they can change over time. \emph{Data snooping}, or repeatedly searching historical data until a favorable result appears, can invalidate conventional tests \cite{lomackinlay1990datasnooping}. Searches across many rules therefore require corrections within families to control false discoveries across the full set \cite{sullivan1999datasnooping,white2000realitycheck}, and the superior predictive ability test addresses comparison against many alternatives \cite{hansen2005spa}. A \emph{backtest} is a simulation on historical data; selecting among many backtests raises the probability of overfitting and motivates the \emph{deflated Sharpe ratio}, which discounts performance for selection and returns that are not normally distributed \cite{bailey2016pbo,bailey2014deflated}. The large set of published return predictors raises the same problem of multiple testing \cite{harvey2016crosssection}, while combining many candidate signals can itself induce severe overfitting \cite{novymarx2015backtesting}. Decay in anomalies after publication reinforces the need for evidence on unseen data and in future deployment \cite{mclean2016academic}. These concerns determine the controls used in our evaluation.

Accordingly, our paired design avoids treating the 100 asset predictions on a date as independent economic observations. We estimate effects on daily portfolio returns with HAC inference \cite{neweywest1987}, disclose all tested contrasts between backbones and information conditions, adjust the three primary tests within each model family, and retain failed outputs under an intention-to-treat rule. The resulting claim is deliberately narrow: we estimate the marginal economic value of reasoning effort inside a frozen financial decision pipeline rather than comparing complete agents or asserting that reasoning has no value in finance.

\begin{table*}[t]
\centering
\caption{Experimental design. Each primary study uses 241 formation dates, meaning trading days when new holdings are selected. Each audit generation is produced by a repeated call on the frozen audit formation dates.}
\label{tab:design}
\small
\begin{tabular}{lllp{4.2cm}rr}
\toprule
Backbone & Baseline reasoning & Additional levels & Information conditions & Formation dates & Audit generations\\
\midrule
DeepSeek-Flash & none & low, high, maximum & numerical; identifiable news; masked news & 241 & 3 per condition\\
GPT-5.6 Luna & none & low & numerical; identifiable news; masked news & 241 & 3 per condition\\
Gemini 3.1 Flash-Lite & minimal & low & numerical; identifiable news; masked news & 241 & 3 per condition\\
\bottomrule
\end{tabular}
\end{table*}

\section{Experimental Design}

\subsection{Research Question and Estimand}

An \emph{estimand} is the population quantity an experiment is designed to measure. For backbone $b$, information condition $a$, reasoning level $r$, and portfolio return date $t$, let $R^{\mathrm{net}}_{b,a,r,t}$ denote the net return of the frozen portfolio implementation. The common low versus baseline estimand is
\begin{equation}
\Delta_{b,a} =
\mathbb{E}\!\left[
R^{\mathrm{net}}_{b,a,\mathrm{low},t}-
R^{\mathrm{net}}_{b,a,\mathrm{base},t}
\right],
\label{eq:estimand}
\end{equation}
where base is none for DeepSeek and GPT and minimal for Gemini. Positive values favor low reasoning. The primary inferential observations are daily portfolio return differences indexed by return date. Asset predictions and model tasks within a formation date are inputs to a single portfolio, not independent return observations.

We predefine the economic threshold $\epsilon=0.25$ bps/day as the smallest improvement considered worthwhile. Using a two-sided 95\% CI $[L,U]$ and a one-sided 95\% upper bound $U_{.95}$, we classify a contrast as: material benefit if $L>\epsilon$; harm if $U<0$; no material benefit if $U_{.95}<\epsilon$; and inconclusive otherwise. This asymmetric rule reflects the deployment question: added reasoning has known compute cost and must demonstrate sufficient incremental value. An inconclusive result is not \emph{equivalence}, which would require ruling out effects outside prespecified small bounds.

\subsection{Market Sample and Information Timing}

The experiment uses liquid common equities listed in the United States. \emph{Information timing} means that each decision uses only information available by its formation date. The initial snapshot, dated December 29, 2023, contains 120 eligible names, of which 100 are active on each evaluated formation date. There are 241 formation dates from January 10 through December 23, 2024. Each formation date's \emph{cross-section}, meaning the 100 stocks compared on that date, is deterministically split into four model tasks of 25 assets each. Assets use \emph{opaque identifiers}, random codes that reveal no company name, and no future return enters an inference packet or repair decision.

These dated decisions become overlapping portfolio observations. A \emph{formation cohort} is the portfolio opened on one formation date. It enters at that session's open, earns open-to-close return on that date, and then close-to-close return on the next four sessions. A \emph{return date} is a trading day on which active formation cohorts earn portfolio returns. After the first four formation dates are removed, the realized path contains 241 return dates, from January 17 through December 30.

The frozen stochastic audit applies the same timing logic to 48 clustered formation dates selected without reference to their eventual returns. Because formation cohorts are held for five days and overlap, the audit portfolio path contains 68 return dates. Repeated generations are averaged or compared within return date; they never multiply the number of market observations.

\subsection{Backbones and Reasoning Levels}

Table~\ref{tab:design} gives the final analyzed panel. All three backbones use numerical, identifiable news, and masked news inputs, but the available reasoning controls differ. DeepSeek tests none, low, high, and maximum reasoning. GPT tests none versus low. Gemini tests minimal versus low because Gemini 3.x could not guarantee a condition with reasoning fully disabled. Only DeepSeek therefore supports a claim about the response across four reasoning levels; GPT and Gemini replicate the common low versus baseline comparison.

The three evaluated backbones were DeepSeek-Flash, GPT-5.6 Luna, and Gemini 3.1 Flash-Lite. We refer to them as DeepSeek, GPT, and Gemini except where the exact model name is needed. All were predetermined for the study. Before each backbone was evaluated, its exact reasoning contrast, inputs, prompts, output contract, failure policy, portfolio mapping, and analysis plan were frozen. Each backbone therefore supplies a separately controlled intervention within one model rather than an exercise in model selection.

\subsection{Information Conditions}

The \emph{numerical} condition supplies 16 daily \emph{cross-sectional rank features}. Each raw measure is ranked across the broader fixed research panel available on that date, scaled to $[-1,1]$, and then retained for the 100 active candidate stocks. The measures are return lagged one day; momentum from past returns over 5, 21, 63, 126, and 252 days; volatility over 5, 21, and 63 days; downside volatility over 21 days; a volume ratio over 21 days; log dollar volume over 21 days; \emph{Amihud illiquidity} over 21 days, the amount of price movement per dollar traded; previous intraday return; previous trading range; and log previous close. For each feature the packet includes its current value, mean over 5 days, mean over 20 days, and change over 20 days.

The news conditions provide article information available by each formation date. \emph{Identifiable news} retains identifying details. \emph{Masked news} removes direct company, ticker, date, and event identifiers while preserving the text structure relevant to the task. This manipulation reduces obvious historical recognition but cannot guarantee that a model has no latent memory of an event.

\subsection{Prompt, Output Contract, and Integrity Controls}

Every model receives the same substantive instruction: estimate relative return over the next five trading sessions, score the strongest candidate near $+1$ and the weakest near $-1$, and return every opaque asset identifier exactly once. The output \emph{schema} is the set of required fields, types, and ranges that a machine can check. It contains \emph{score} in $[-1,1]$, \emph{confidence}, \emph{materiality}, and \emph{uncertainty} in $[0,1]$, plus an \emph{abstain} flag that lets the model decline to score an asset. Prompts, asset packets, candidate order, schema, output allowance, and downstream mapping are frozen within each paired comparison.

Responses undergo schema and identifier validation. Retries and repairs under deterministic rules are \emph{outcome blind}, meaning they are made without viewing subsequent returns; raw attempts, usage, and repair records are retained. When a task remains invalid, all 25 rows are assigned a neutral score of 0 and kept under the primary intention-to-treat rule. DeepSeek ended with complete task coverage after 17 deterministic contract repairs, affecting 15 of 404,400 asset rows by neutral imputation. Across the three primary information conditions, GPT retained 57 invalid tasks (0.99\%) and Gemini retained 136 (2.35\%). Analyses that remove incomplete dates or match omitted task chunks are reported as robustness checks, not replacements for the frozen primary analysis.

\begin{figure*}[!t]
\centering
\begin{tikzpicture}
\begin{axis}[
  width=0.93\textwidth,height=7.4cm,
  xmin=-11,xmax=21,
  xlabel={Effect of low relative to baseline reasoning on net return (bps/day)},
  ytick={1,2,3,4,5,6,7,8,9},
  yticklabels={Gemini masked,Gemini identifiable,Gemini numerical,GPT masked,GPT identifiable,GPT numerical,DeepSeek masked,DeepSeek identifiable,DeepSeek numerical},
  y dir=normal,
  grid=both,
  major grid style={draw=gray!20},
  minor grid style={draw=gray!10},
  minor x tick num=1,
  axis y line*=left,
  axis x line*=bottom,
  tick label style={font=\footnotesize},
  label style={font=\footnotesize},
  clip=false]
\addplot[black,line width=1.4pt] coordinates {(0,0.5) (0,9.5)};
\addplot[black,dashed,line width=1.0pt] coordinates {(0.25,0.5) (0.25,9.5)};
\node[anchor=south west,font=\scriptsize] at (axis cs:0.45,9.20) {$+0.25$ economic threshold};
\addplot[only marks,solid,color=deepseek,mark=*,mark size=3pt,mark options={fill=deepseek,draw=deepseek},line width=1.2pt,error bars/.cd,x dir=both,x explicit,error bar style={solid,line width=1.2pt},error mark options={solid,rotate=90,mark size=3pt,line width=1.2pt}]
  coordinates {(6.159,9) +- (13.402,0)};
\addplot[only marks,solid,color=deepseek,mark=square*,mark size=3pt,mark options={fill=deepseek,draw=deepseek},line width=1.2pt,error bars/.cd,x dir=both,x explicit,error bar style={solid,line width=1.2pt},error mark options={solid,rotate=90,mark size=3pt,line width=1.2pt}]
  coordinates {(0.378,8) +- (9.411,0)};
\addplot[only marks,solid,color=deepseek,mark=diamond*,mark size=3.5pt,mark options={fill=deepseek,draw=deepseek},line width=1.2pt,error bars/.cd,x dir=both,x explicit,error bar style={solid,line width=1.2pt},error mark options={solid,rotate=90,mark size=3pt,line width=1.2pt}]
  coordinates {(1.449,7) +- (8.217,0)};
\addplot[only marks,solid,color=gpt,mark=*,mark size=3pt,mark options={fill=gpt,draw=gpt},line width=1.2pt,error bars/.cd,x dir=both,x explicit,error bar style={solid,line width=1.2pt},error mark options={solid,rotate=90,mark size=3pt,line width=1.2pt}]
  coordinates {(3.540,6) +- (6.071,0)};
\addplot[only marks,solid,color=gpt,mark=square*,mark size=3pt,mark options={fill=gpt,draw=gpt},line width=1.2pt,error bars/.cd,x dir=both,x explicit,error bar style={solid,line width=1.2pt},error mark options={solid,rotate=90,mark size=3pt,line width=1.2pt}]
  coordinates {(-0.626,5) +- (5.279,0)};
\addplot[only marks,solid,color=gpt,mark=diamond*,mark size=3.5pt,mark options={fill=gpt,draw=gpt},line width=1.2pt,error bars/.cd,x dir=both,x explicit,error bar style={solid,line width=1.2pt},error mark options={solid,rotate=90,mark size=3pt,line width=1.2pt}]
  coordinates {(0.040,4) +- (4.338,0)};
\addplot[only marks,solid,color=gemini,mark=*,mark size=3pt,mark options={fill=gemini,draw=gemini},line width=1.2pt,error bars/.cd,x dir=both,x explicit,error bar style={solid,line width=1.2pt},error mark options={solid,rotate=90,mark size=3pt,line width=1.2pt}]
  coordinates {(0.758,3) +- (6.991,0)};
\addplot[only marks,solid,color=gemini,mark=square*,mark size=3pt,mark options={fill=gemini,draw=gemini},line width=1.2pt,error bars/.cd,x dir=both,x explicit,error bar style={solid,line width=1.2pt},error mark options={solid,rotate=90,mark size=3pt,line width=1.2pt}]
  coordinates {(-0.142,2) +- (7.986,0)};
\addplot[only marks,solid,color=gemini,mark=diamond*,mark size=3.5pt,mark options={fill=gemini,draw=gemini},line width=1.2pt,error bars/.cd,x dir=both,x explicit,error bar style={solid,line width=1.2pt},error mark options={solid,rotate=90,mark size=3pt,line width=1.2pt}]
  coordinates {(-1.123,1) +- (7.296,0)};
\end{axis}
\end{tikzpicture}
\caption{Primary low versus baseline return effects across 241 return dates. Circles denote numerical inputs, squares denote identifiable news, and diamonds denote masked news. Horizontal bars are two-sided 95\% HAC CIs, which account for changing variance and serial dependence. The solid line is zero and the dashed line is the $+0.25$ bps/day economic threshold. No interval establishes a material benefit.}
\label{fig:forest}
\Description{A horizontal forest plot shows nine effects of low reasoning relative to baseline across numerical, identifiable news, and masked news inputs. All confidence intervals cross zero and the positive economic threshold.}
\end{figure*}
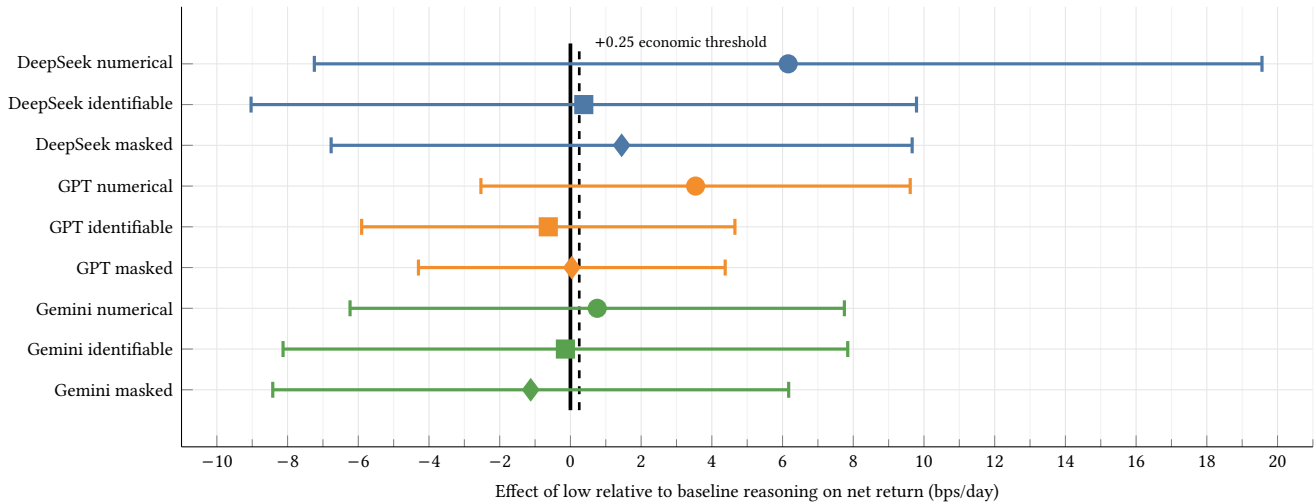

\section{Portfolio Evaluation and Statistical Inference}

\subsection{Mapping Scores to Portfolios}

The portfolio mapping converts each model output into one comparable trading signal. For asset $i$ on formation date $d$, the downstream signal is
\begin{equation}
s_{i,d}=\begin{cases}
0, & \text{if abstaining or imputed to }0,\\
q_{i,d}\left(0.5+0.5c_{i,d}\right), & \text{otherwise}.
\end{cases}
\label{eq:signal}
\end{equation}
Here, $q$ is the model score and $c$ is model reported confidence. The model reported \emph{materiality} and \emph{uncertainty} fields are preserved for diagnostics but do not alter position size.

We then rank $s_{i,d}$ across the 100 names on each formation date, short the bottom 10, and buy the top 10. Positions receive equal weights within each side. Thus \emph{gross exposure} is 2 (one dollar long plus one dollar short per dollar of capital) and \emph{net exposure} is 0 (long and short dollar amounts cancel). If abstentions leave fewer than 20 eligible assets, the frozen fallback ranks all neutralized signals and breaks ties deterministically by asset identifier.

Each formation cohort is held for five sessions. The implemented daily portfolio is the simple average of active cohorts, so formation decisions overlap in calendar time. All reasoning levels use the identical simulator and candidate universe.

\subsection{Trading and Inference Costs}

The primary assumption for trading costs is 10 bps one way, with 5, 20, and 50 bps sensitivity analyses. For a portfolio with capital $A$, incremental inference expense $C$ can be expressed as
\begin{equation}
\text{inference cost (bps)}=10{,}000\,C/A.
\label{eq:inferencecost}
\end{equation}
We keep provider inference dollars separate from market returns because the same research workload can support different capital levels. Deployment conclusions use the direction and uncertainty of the effect after trading costs first, then ask whether any supported gain could cover inference expense.

\subsection{Paired Inference}

For every comparison between a backbone and information condition, we align the two portfolio paths by return date and compute $D_t=R^{\mathrm{net}}_{\mathrm{low},t}-R^{\mathrm{net}}_{\mathrm{base},t}$. The mean and \emph{standard error}, an estimate of sampling variability, use a Newey--West estimator with five lags to account for changing variance and the serial dependence induced by overlapping formation cohorts \cite{neweywest1987}. For the comparable annual low versus baseline results, all three backbones have three information conditions. Their two-sided \emph{$p$-values}, which give the probability under an assumption of zero effect of obtaining a result at least this extreme, are adjusted with the \emph{Holm method}, a stepwise correction that controls false positives across the three tests within each backbone. This harmonized reporting layer does not replace DeepSeek's predeclared sequential procedure: none to low, low to maximum when the first comparison is unresolved, and the total effect from none to maximum. The collected high level is analyzed as part of a secondary curve across four reasoning levels.

As supplementary uncertainty checks, we add two \emph{bootstraps}, which estimate variability by repeatedly resampling the observed data. They do not replace the primary HAC inference. A paired \emph{moving block bootstrap} \cite{kunsch1989bootstrap} resamples 50,000 sequences of consecutive return dates. Its main block of five sessions matches the holding horizon, while blocks of 10 and 20 sessions test sensitivity. A \emph{pairs bootstrap over formation cohorts} instead resamples each cohort's complete contribution to gross returns over five sessions together with its paired difference in transaction costs. The cohort contributions reconcile exactly to the observed sum of net returns. Confidence intervals use the 2.5th and 97.5th percentiles of the resampled effects, and two-sided tests center the resampled differences at zero. The moving block result is the stronger check of dependence over calendar time because the cohort bootstrap treats formation cohorts as interchangeable even though nearby cohorts can share shocks on the same return date.

Together, these procedures produce the point estimates, two-sided CIs, one-sided economic bounds, and decisions reported in the paper. A conventional \emph{null hypothesis test} begins from a default assumption of zero effect; merely failing to reject that assumption is not treated as evidence that reasoning has no effect.

We next ask whether an annual estimate is concentrated in one part of the year. As an exploratory diagnostic, we partition each saved primary daily effect series by return date into calendar halves (January--June and July--December) and quarters. We recompute HAC intervals with five lags within each subgroup and use HAC regressions for \emph{heterogeneity tests}, which ask whether the effect differs across halves or quarters. We adjust the resulting 18 tests with the Holm method. We also remove each quarter in turn and measure its share of the sum of absolute quarterly contributions. These partitions use no new model calls, do not change the annual estimands, and are not treated as independent replications.

\subsection{Evaluation Across Stochastic Generations}

On the frozen audit formation dates we assess three layers of reproducibility: (i) \emph{Spearman rank agreement}, which measures whether assets keep a similar ordering; (ii) \emph{Jaccard overlap}, the fraction of names shared by the selected long and short sets; and (iii) the difference between low and baseline reasoning for each generation. For every information condition, each backbone has the primary generation restricted to the audit dates plus two additional generations. For GPT and Gemini we also form a portfolio from \emph{ensemble scores} by averaging the three signals before ranking, alongside the average within each return date of the three return effects. Because ranking and selection of the sets with the highest and lowest ranks are nonlinear, the ensemble score effect need not equal the arithmetic mean of the three return effects.

\section{Results}

\subsection{RQ1: Does Low Reasoning Improve Annual Performance?}

Figure~\ref{fig:forest} and Table~\ref{tab:primary} report the nine comparable annual low versus baseline effects. None satisfies the rule for material benefit. Six point estimates are positive and three are negative, but every 95\% CI includes zero and the $+0.25$ bps/day economic threshold. DeepSeek's numerical point estimate is the largest at $+6.159$ bps/day, yet its interval ranges from $-7.243$ to $+19.560$. The identifiable news estimates are $+0.378$ for DeepSeek, $-0.626$ for GPT, and $-0.142$ for Gemini, and all three intervals likewise admit gains and losses. Holm adjustment across the three information conditions within each backbone does not change any decision. The annual evidence therefore supports \emph{no reliable benefit detected}, not the stronger claim of a zero effect.

\begin{table*}[t]
\centering
\caption{Primary low versus baseline results at a one way trading cost of 10 bps. Adjusted $p$-values use the Holm method across the three information conditions separately within each backbone.}
\label{tab:primary}
\small
\begin{tabular}{llrrrrl}
\toprule
Backbone & Information & Estimate & 95\% CI & $p$ & Adjusted $p$ & Decision\\
\midrule
DeepSeek & numerical & $+6.159$ & $[-7.243,19.560]$ & .368 & 1.000 & inconclusive\\
DeepSeek & identifiable news & $+0.378$ & $[-9.033,9.790]$ & .937 & 1.000 & inconclusive\\
DeepSeek & masked news & $+1.449$ & $[-6.768,9.666]$ & .730 & 1.000 & inconclusive\\
GPT & numerical & $+3.540$ & $[-2.531,9.611]$ & .253 & .759 & inconclusive\\
GPT & identifiable news & $-0.626$ & $[-5.905,4.653]$ & .816 & 1.000 & inconclusive\\
GPT & masked news & $+0.040$ & $[-4.297,4.378]$ & .986 & 1.000 & inconclusive\\
Gemini & numerical & $+0.758$ & $[-6.233,7.749]$ & .832 & 1.000 & inconclusive\\
Gemini & identifiable news & $-0.142$ & $[-8.128,7.844]$ & .972 & 1.000 & inconclusive\\
Gemini & masked news & $-1.123$ & $[-8.418,6.173]$ & .763 & 1.000 & inconclusive\\
\bottomrule
\end{tabular}
\end{table*}

These paired effects describe the value of additional reasoning, not the absolute attractiveness of either portfolio. Appendix Table~\ref{tab:absolute-performance} reports the underlying paths: 13 of the 18 portfolios at the baseline and low reasoning levels have negative mean net return. The positive annualized returns are confined to Gemini identifiable news at both levels and Gemini masked news at minimal reasoning; the largest is $2.16\%$. A \emph{compounded annualized return} is the equivalent annual growth rate. A \emph{Sharpe ratio} divides mean return by return volatility, and \emph{maximum drawdown} is the largest portfolio loss from a peak to a trough.

The conclusion also depends on the frozen portfolio construction. Appendix Table~\ref{tab:tail-sensitivity} varies the number of selected names. Across 36 exploratory combinations of backbone, information condition, and width, 35 do not establish material benefit. The exception is DeepSeek masked news with 20 names per side ($+5.769$ bps/day; HAC 95\% CI $[0.597,10.942]$), which also has a positive interval from a moving block bootstrap with five sessions per block. However, no HAC or moving block $p$-value survives Holm correction across the 36 cells. The exception does not alter the frozen primary estimand based on 10 long and 10 short names, but it bounds the conclusion to that tested construction.

\subsection{RQ2: Does More Reasoning Consistently Improve Returns?}

Only DeepSeek tests a curve across four reasoning levels. Figure~\ref{fig:level-stability}a is a signed matrix of point estimates averaged across generations for portfolios formed on the audit sample of 48 formation dates, relative to none. The printed values make the nonmonotonic patterns explicit without implying a smooth response between provider levels. For numerical inputs, the effect declines from none through low, high, and maximum. For identifiable news, low reasoning drops sharply, high changes little, and maximum recovers part of the loss but not all of it. Masked news remains below none at every positive reasoning level. These are descriptive level estimates; confidence intervals are attached to the predeclared contrasts rather than invented for cumulative points.

The confirmatory chain is consistent with the visual result. Low minus none is $-0.529$ bps/day for numerical inputs, $-8.525$ for identifiable news, and $-9.016$ for masked news. The CI for masked news, $[-16.248,-1.784]$, establishes harm. Maximum minus none remains inconclusive for numerical and identifiable news and is classified as no material benefit for masked news. The only decisive adjacent secondary contrast is maximum minus high for identifiable news, $+4.639$ bps/day (95\% CI $[+0.376,+8.903]$). Because maximum still fails to beat none, a favorable local step does not justify the claim that more reasoning is better.

\subsection{RQ3: Are Reasoning Effects Stable Across Generations?}

\begin{figure*}[!t]
\centering
\includegraphics[width=0.94\textwidth]{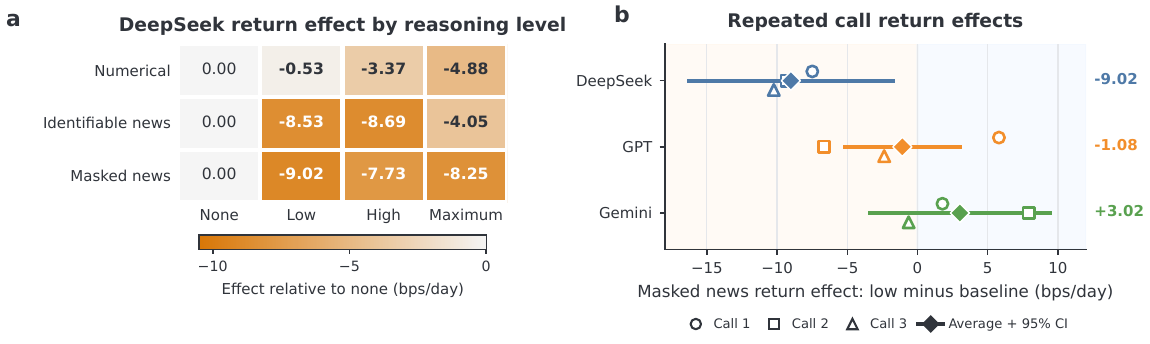}
\caption{Reasoning levels and repeated call stability. (a) DeepSeek effects averaged across generations for portfolios formed on 48 audit formation dates, relative to none; cells are descriptive because intervals for cumulative levels were not computed. (b) Three unconnected generation markers per backbone; diamonds and bars show the average within each return date and the 95\% HAC CI adjusted for serial dependence. Repeated calls are not independent market samples.}
\label{fig:level-stability}
\Description{Panel a is a signed matrix of DeepSeek effects at none, low, high, and maximum reasoning for numerical, identifiable news, and masked news inputs. Panel b plots three separate generation markers for masked news and an aggregate confidence interval for each backbone. DeepSeek is consistently negative, while GPT and Gemini each change sign.}
\end{figure*}

Figure~\ref{fig:level-stability}b begins with masked news, where the contrast between reproducible harm and unstable direction is clearest. It shows low versus baseline estimates as three unconnected generation markers for each backbone, together with the average within each return date and its 95\% HAC interval. DeepSeek's three individual audit generations are all negative ($-7.490$, $-9.327$, and $-10.231$ bps/day), producing the harm result after aggregation within each return date. GPT instead changes sign across $+5.800$, $-6.665$, and $-2.371$ bps/day. Its average across three generations within each return date is $-1.079$ bps/day (95\% CI $[-5.173,3.015]$), while its ensemble score portfolio is likewise inconclusive at $-0.957$ bps/day (95\% CI $[-6.814,4.900]$).

Gemini also changes sign: its first two generations are positive ($+1.777$ and $+7.918$ bps/day), whereas the third is $-0.630$ bps/day. The average within each return date is $+3.022$ bps/day (95\% CI $[-3.377,9.421]$), and the ensemble score portfolio remains inconclusive at $+1.609$ bps/day (95\% CI $[-6.545,9.764]$). Thus the stability result is not that all generations are negative. It is that a favorable isolated generation does not establish reproducible economic value. The third generations retain one of 384 GPT tasks and 13 of 384 Gemini tasks as neutral after invalid outputs; none is selectively rerun. These audits do not replace the primary estimates from 241 return dates or add independent market observations.

The other information conditions preserve the deployment concern without reproducing the same directions. For identifiable news, GPT's generation effects are $+0.082$, $-0.649$, and $-3.881$ bps/day. Their average within each return date is $-1.483$ bps/day, with a 95\% CI from $-5.940$ to $+2.974$, while the score ensemble is $-3.604$ bps/day, with a CI from $-8.699$ to $+1.491$. Gemini changes sign across $-0.688$, $+1.476$, and $+8.870$ bps/day. Its average within each return date is $+3.219$ bps/day, with a CI from $-7.032$ to $+13.470$, and its ensemble is $+3.882$ bps/day, with a CI from $-6.561$ to $+14.325$. None establishes a reliable benefit. Appendix Table~\ref{tab:stochastic-agreement} shows that similar overall scores still yield incomplete agreement in the selected portfolio tails.

Numerical inputs complete the same three generation design. GPT's effects are $-3.972$, $-0.927$, and $-1.112$ bps/day. Their average within each return date is $-2.004$ bps/day (95\% CI $[-6.386,2.379]$), and the score ensemble is $-3.866$ bps/day (95\% CI $[-9.870,2.138]$). Gemini's generation effects are $+2.750$, $+0.788$, and $+2.290$ bps/day. Their average within each return date is $+1.943$ bps/day (95\% CI $[-6.161,10.047]$), while the score ensemble is $-0.222$ bps/day (95\% CI $[-8.158,7.714]$). This sign reversal after score averaging illustrates that stable looking scores do not guarantee a stable selected portfolio. None of the numerical audit summaries establishes a reliable benefit. The two added GPT generations retain six of 768 tasks as neutral after invalid outputs; both added Gemini generations have complete valid coverage.

\subsection{Robustness Across Resampling and Calendar Periods}

We first test whether the conclusions depend on the primary HAC calculation. The resampling checks preserve the headline decisions (Appendix Table~\ref{tab:bootstrap}). All nine low versus baseline contrasts across 241 return dates remain inconclusive under both the moving block bootstrap with five sessions per block and the bootstrap over formation cohorts. For identifiable news, the moving block intervals are $[-9.099,9.287]$ for DeepSeek, $[-6.020,4.586]$ for GPT, and $[-7.690,8.497]$ bps/day for Gemini. Changing the block length to 10 or 20 sessions does not produce a decision of material benefit. In the DeepSeek audit of three generations using masked news, the moving block interval is $[-17.929,-3.477]$ and the interval from the formation cohort bootstrap is $[-15.482,-2.281]$ bps/day, so the harm classification survives both checks.

We then test whether the annual directions are confined to one part of 2024 (Appendix Table~\ref{tab:calendar-subperiods}). Only GPT numerical keeps the same sign in both halves of the year. DeepSeek numerical changes from $-10.027$ bps/day in the first half to $+20.687$ in the second (raw HAC heterogeneity $p=.014$), while Gemini masked news ranges from $+12.083$ in Q3 to $-17.974$ in Q4 (raw test of equality across quarters $p=.045$). Neither result survives Holm correction across the 18 exploratory heterogeneity tests by half and quarter. Seven of nine annual directions also flip when at least one quarter is omitted, although the largest absolute quarter contribution is at most 64.5\%. The annual findings are therefore not reducible to one isolated quarter, but neither are they temporally uniform. This analysis strengthens the conclusion that reasoning effects vary across market episodes within 2024; it does not substitute for another year.

\section{Operational Reliability and Compute Economics}

\subsection{Output Structure Reliability}

Reasoning affects task validity differently across backbones. Fisher's exact test, which compares two failure proportions, is used as a diagnostic. GPT low reasoning increases numerical failures from 4 to 14 of 964 tasks (0.41\% to 1.45\%; $p=.030$), while failures for masked news are similar (8 versus 9). For identifiable news, GPT failures fall from 17 to 5 ($p=.016$), and Gemini failures fall from 54 to 14 ($p<.000001$). Gemini low reasoning also reduces failures for masked news from 44 to 20 of 964 tasks (4.56\% to 2.07\%; $p=.0032$), but neither reliability improvement produces a return benefit. DeepSeek has no final invalid tasks after rare outcome blind repairs. Reliability is therefore a separate deployment dimension, not a proxy for investment performance; Table~\ref{tab:compute} retains the primary rates of invalid tasks.

\subsection{Tokens, Dollars, and the Deployment Decision}

Because reliability and return can diverge, deployment also requires a separate accounting of compute. Table~\ref{tab:compute} reports compute, cost, and reliability for all evaluated tasks within each backbone. These are backbone totals rather than efficiency comparisons at a matched workload. DeepSeek dominates compute because it evaluates four reasoning levels across all three information conditions. Its evaluation uses 269.0 million reported reasoning tokens. GPT uses 0.37 million reasoning tokens, while Gemini uses 7.93 million, illustrating that nominal level labels are not comparable measures of actual compute across providers.

For GPT's primary evaluation, low reasoning costs an additional \$0.375 for numerical inputs and \$0.092 for masked news, while identifiable news costs \$0.281 less because the low level produces fewer output tokens in this run. These provider charges are modest relative to the return intervals, but a low dollar cost does not turn an unsupported return gain into evidence of benefit. Latency also depends on the provider and delivery mode and is not used as a comparable headline metric.

\begin{table*}[t]
\centering
\caption{Compute, cost, and reliability for all evaluated tasks within each backbone. Token counts are reported by providers and should not be compared as identical units of cognition. Invalid primary tasks is the share of primary tasks still unresolved after the frozen validation and repair policy. NA denotes not aggregated; token columns are not aggregated across providers, and total cost is summed before row values are rounded.}
\label{tab:compute}
\small
\begin{tabular}{lrrrrrr}
\toprule
Backbone & Tasks & Prompt tokens & Completion tokens & Reasoning tokens & Cost (USD) & Invalid tasks\\
\midrule
DeepSeek-Flash & 16,176 & 801,697,928 & 288,221,145 & 268,969,372 & 297.52 & 0.00\%\\
GPT-5.6 Luna & 8,088 & 360,029,538 & 11,992,166 & 371,323 & 41.98 & 0.99\%\\
Gemini 3.1 Flash-Lite & 8,088 & 507,532,598 & 17,670,460 & 7,928,611 & 75.56 & 2.35\%\\
\midrule
Total & 32,352 & NA & NA & NA & 415.05 & NA\\
\bottomrule
\end{tabular}
\end{table*}

\section{Discussion}

\subsection{Reasoning Is Not a Monotonic Economic Control}

The results distinguish behavioral responsiveness from economic usefulness. Reasoning effort clearly changes outputs: token use rises, rankings move, portfolio constituents change, and rates of invalid outputs sometimes shift. Yet those changes do not form a stable monotonic relationship with returns after costs. DeepSeek provides the strongest demonstration. An adjacent increase from high to maximum helps identifiable news on the audit sample, but maximum still fails to outperform none. Treating reasoning as a conventional scalar \emph{hyperparameter}, or a tunable numerical setting for which more is often presumed better, is therefore misleading in this setting.

This pattern is compatible with several mechanisms, none of which our experiment proves. Longer traces may assign too much weight to weak narrative evidence, revise a useful first judgment, increase confidence without increasing information, or simply create another stochastic route through an underdetermined task. Conversely, reasoning may improve local extraction or schema compliance while leaving the economically relevant rank order unchanged. A causal mechanism study would require randomized interventions on trace content or a task with intermediate labels, not inspection of provider summaries alone.

\subsection{What the Negative Result Does and Does Not Mean}

The most defensible statement is that additional low reasoning did not deliver a reliable improvement in this frozen 2024 pipeline. It is stronger than saying that nine conventional $p$-values exceed 0.05 because it incorporates transaction costs, an economic threshold, repeated generations, and operational failures. It is weaker than equivalence, which requires formal evidence that effects lie within small prespecified bounds, because most CIs still include economically meaningful gains and losses. The study also does not establish that the baseline trading strategies are profitable, that reasoning cannot help other financial tasks, or that the evaluated provider controls correspond to the same quantity of compute.

\subsection{Implication for Governed Deployment}

In an institutional process for model risk, increasing reasoning effort should be treated as a configuration change that requires downstream validation, not as a capability upgrade that is beneficial by default. Approval evidence should match the deployed decision rule and include action stability, failures, costs, and performance across repeated generations. Under that standard, an isolated favorable generation or an unresolved positive point estimate is insufficient to promote a more expensive reasoning level. The conservative default in this study is therefore a governance decision under uncertainty, not a claim that baseline reasoning is intrinsically superior.

\section{Limitations}

The first limitation is temporal and model scope. Evidence comes from one 2024 backtest of U.S. equities and the provider interfaces observed at execution. The analysis by calendar period spans distinct episodes within the year and shows that no contrast is wholly driven by one quarter, but the frequent sign changes reinforce the need for replication across years or in a prospective study. Masking reduces cues to recognizable events but cannot rule out latent historical knowledge, and future provider updates may change behavior. Reasoning level support is also asymmetric: only DeepSeek exposes the four tested levels, so provider labels should not be interpreted as equal units of compute.

The second limitation is precision and portfolio dependence. Most primary intervals admit economically meaningful gains and losses; the 80\% minimum detectable effects are 6.200--19.156 bps/day versus the 0.25 bps/day economic threshold (Appendix Table~\ref{tab:detectable-effects}). The result is therefore a conservative deployment decision, not an equivalence claim. It is also tied to the frozen rule that selects 10 long and 10 short names: one of 36 exploratory width cells is positive before, but not after, correction across the family.

The final limitation concerns deployment and mechanism. The simulator models fixed transaction costs and overlapping holdings but not nonlinear \emph{market impact} (prices moving because of the strategy's own trades), borrow availability for short sales, capacity, latency, or uncertainty in live orders. The experiment estimates the total effect of requesting more reasoning; it does not identify the hidden computation causing a score or trade change. These omissions lead directly to the next validation step: prospective \emph{shadow trading}, which runs the strategy on live inputs without placing orders, is required before any claim about live deployment.

\section{Conclusion}

We evaluate inference-time reasoning as an economic intervention in a frozen LLM system for ranking equities. Across DeepSeek, GPT, and Gemini, none of nine low versus baseline tests across 241 return dates establishes a material improvement in net portfolio return. DeepSeek's replicated result with masked news shows harm, its curve across four reasoning levels is nonmonotonic, and repeated generations reveal sign instability across the news conditions. Reasoning can also improve or degrade output reliability without improving returns. In this setting, extra thought changes decisions but does not reliably pay for itself. For the primary implementation with 10 long and 10 short names, baseline reasoning is therefore the appropriate default until a prospective study of the target task demonstrates reproducible incremental value. The exploratory exception for portfolio width shows why that recommendation should not be transferred unchanged to a different portfolio rule.

\clearpage
\bibliographystyle{ACM-Reference-Format}
\begingroup
\setlength{\bibsep}{-0.75pt}
\bibliography{references}
\endgroup

\clearpage
\appendix
\section{Prompt and Structured Output Contract}

The system instruction casts the model as a conservative cross-sectional equity forecaster, treats supplied article text as untrusted data, restricts the response to the supplied inputs, and requires exactly one structured text object in JavaScript Object Notation (JSON). The user prompt asks for relative return over the next five trading sessions. It instructs the model to favor persistent evidence across horizons and penalize reversal, unstable volatility, illiquidity, contradiction, and uncertainty. In news tasks it distinguishes new material information about the target from repetition, commentary, price narration, legal advertising, and noise.

Each of the 25 expected opaque identifiers must appear exactly once in:
\begin{verbatim}
{"items":[
  {"asset":"A0123456789",
   "score":0.0,
   "confidence":0.0,
   "materiality":0.0,
   "uncertainty":0.0,
   "abstain":false}
]}
\end{verbatim}
The exact prompts are stored as compressed immutable artifacts with hashes. Schema syntax differs by provider, but the fields and bounds are common.

\subsection{Reasoning Intervention Settings}

\begin{table}[h]
\centering
\caption{Provider controls defining the reasoning intervention. Prompt, inputs, schema, candidate order, and output limit remain fixed within each backbone.}
\label{tab:reasoning-settings}
\small
\resizebox{\columnwidth}{!}{%
\begin{tabular}{@{}llllr@{}}
\toprule
Backbone & Baseline & Added reasoning & Provider control & Output limit\\
\midrule
DeepSeek & none & low, high, maximum & \texttt{thinking}; \texttt{reasoning\_effort} & 65,536\\
GPT & none & low & \texttt{reasoning.effort} & 45,000\\
Gemini & minimal & low & \texttt{thinkingLevel} & 8,192\\
\bottomrule
\end{tabular}}
\end{table}

DeepSeek disables thinking for none and enables it with the requested effort otherwise; GPT varies \texttt{reasoning.effort}; Gemini varies \texttt{thinkingLevel}. No condition receives an additional reasoning instruction. Provider labels and output limits are not directly comparable quantities of computation. Reasoning summaries or traces are retained for auditing but are not treated as faithful records of hidden computation.

\subsection{Numerical Data Provenance and Volume}

The numerical condition is derived from the Center for Research in Security Prices (CRSP) daily stock files accessed through WRDS. The source fields include daily prices, returns, trading volume, opening prices, high and low price fields, share codes, exchange codes, and delisting records. Dated security name intervals restrict the sample to ordinary common shares listed on the NYSE, AMEX, or Nasdaq. Reported returns are combined with delisting returns. Under the frozen cleaning rule, when a delisting code from 500 to 599 has no reported delisting return, the return is set to $-30\%$; these codes indicate a delisting related to poor performance. Other missing delisting returns remain missing. All model features end at the preceding trading session's close.

Raw CRSP files from 2020 through 2024 support the required histories. The resulting feature construction panel fixes 1,000 liquid securities using information ending December 30, 2022, and contains 974,641 security and date rows over 1,005 trading dates from January 4, 2021, through December 31, 2024. Feature ranks are computed within this broader panel before values are retained for the 100 active candidates. Across the 241 formation dates in the experiment, the models receive 24,100 stock and date numerical records. Each record contains 16 ranked features and four summaries per feature, for 64 values per stock and 1,542,400 values in the unique 2024 numerical input panel. The same records are reused across reasoning levels and model families, so repeated calls do not create additional market observations.

\subsection{News Data Provenance and Volume}

The news conditions use the licensed EOD Historical Data (EODHD) news feed. Articles were linked to the active CRSP security universe through unambiguous provider ticker mappings. For each formation date, an article was eligible only if its provider timestamp fell from 4:00 p.m. Eastern Time on the preceding trading session up to, but not including, 9:00 a.m. Eastern Time on the formation date. Duplicate query returns were normalized to stable article identifiers, and the three most recent eligible articles for each stock and formation date were retained. Each model received the publisher, title, elapsed hours before the decision, and full article text, subject to fixed context limits of 9,000 characters per article and 18,000 characters per stock. Identifiable and masked news used the same articles. Masking replaced issuer names, tickers, explicit calendar dates and years, and the publisher label while otherwise preserving the supplied content.

Across 241 formation dates and 100 stocks per date, 16,916 of 24,100 stock and date combinations (70.2\%) contained at least one eligible article, while 7,184 (29.8\%) contained none. The retained packets contained 38,097 article references linked to stocks, representing 29,444 unique provider article identifiers and 13 publisher labels. Among the 16,916 nonempty packets, 4,623 contained one article, 3,405 contained two, and 8,888 contained three. A source article linked to more than one stock is counted once among unique identifiers but once for each stock among the 38,097 references.

\section{Complete DeepSeek Confirmatory Chain}

\begin{table}[h]
\centering
\caption{DeepSeek audit across three generations, bps/day. Return date remains the market sampling dimension.}
\small
\resizebox{\columnwidth}{!}{%
\begin{tabular}{lrrr}
\toprule
Condition & Low--none & Maximum--low & Maximum--none\\
\midrule
Numerical & $-0.529$ $[-16.954,15.896]$ & $-4.346$ $[-8.979,0.286]$ & $-4.875$ $[-20.319,10.569]$\\
Identifiable news & $-8.525$ $[-18.208,1.158]$ & $+4.472$ $[-1.595,10.538]$ & $-4.053$ $[-12.949,4.843]$\\
Masked news & $-9.016$ $[-16.248,-1.784]$ & $+0.770$ $[-4.812,6.352]$ & $-8.246$ $[-17.641,1.149]$\\
\bottomrule
\end{tabular}}
\end{table}

The secondary adjacent contrasts (point estimates in bps/day) are: numerical, low--none $-0.529$, high--low $-2.839$, maximum--high $-1.507$; identifiable news, $-8.525$, $-0.167$, $+4.639$; and masked news, $-9.016$, $+1.289$, $-0.519$. Only maximum--high for identifiable news has a two-sided 95\% CI entirely above zero.

\section{Sensitivity to Failure Handling}

GPT's 57 invalid primary tasks and Gemini's 136 invalid primary tasks remain imputed as neutral in the intention-to-treat estimate. Two frozen sensitivity analyses test this choice. The \emph{complete date analysis} removes every formation date containing any failure. The \emph{matched chunk analysis} removes the same chunk of 25 assets from both reasoning levels whenever either level fails. For both GPT and Gemini, all three information conditions remain inconclusive under both analyses. Thus failure handling is operationally important but does not determine the headline return conclusion.

DeepSeek's deterministic repair procedure affected 17 of 16,176 tasks. Across the backbone evaluation, 15 omitted asset rows were imputed as neutral and seven asset identifier errors with an edit distance of one were corrected when the intended match was unique. Repairs were completed without reading outcomes, and the raw attempts and amendments are retained.

\section{Bootstrap Robustness}

\begin{table}[h]
\centering
\caption{Paired bootstrap intervals at a one way cost of 10 bps. MBB-5 is the moving-block bootstrap with five sessions per block; cohort resamples complete formation contributions. Each uses 50,000 deterministic resamples.}
\label{tab:bootstrap}
\scriptsize
\resizebox{\columnwidth}{!}{%
\begin{tabular}{lrr}
\toprule
Contrast & MBB-5 95\% CI & Cohort 95\% CI\\
\midrule
DeepSeek numerical & $[-7.264,20.328]$ & $[-2.958,15.670]$\\
DeepSeek identifiable news & $[-9.099,9.287]$ & $[-7.140,7.849]$\\
DeepSeek masked news & $[-6.635,9.764]$ & $[-5.783,8.715]$\\
GPT numerical & $[-1.832,10.338]$ & $[-2.104,9.436]$\\
GPT identifiable news & $[-6.020,4.586]$ & $[-6.269,5.015]$\\
GPT masked news & $[-4.093,4.667]$ & $[-4.767,4.919]$\\
Gemini numerical & $[-6.933,7.330]$ & $[-6.635,8.097]$\\
Gemini identifiable news & $[-7.690,8.497]$ & $[-9.419,9.555]$\\
Gemini masked news & $[-8.423,6.766]$ & $[-8.127,5.940]$\\
DeepSeek masked news, audit generations 1--3 & $[-17.929,-3.477]$ & $[-15.482,-2.281]$\\
\bottomrule
\end{tabular}}
\end{table}

These supplementary checks do not replace the primary HAC analysis. All nine annual contrasts remain inconclusive, and the audit contrast remains harmful. Moving-block sensitivities at 10 and 20 sessions also leave every annual contrast inconclusive and the audit harmful. Draws in a machine readable format, cluster contributions, exact centered bootstrap $p$-values, and deterministic seeds are retained in the robustness artifact.

\section{Sensitivity to Transaction Costs}

The portfolio simulator replays every condition at one way costs of 5, 10, 20, and 50 bps. GPT's low versus baseline effects remain inconclusive at every cost. Its numerical estimate ranges from $+3.530$ bps/day at 5 bps cost to $+3.619$ at 50 bps; its identifiable news estimate ranges from $-0.647$ to $-0.460$; and its estimate for masked news ranges from $+0.004$ to $+0.332$. Gemini likewise remains inconclusive: numerical effects range from $+0.721$ to $+0.762$, identifiable news ranges from $-0.060$ to $-0.803$, and masked news ranges from $-1.574$ to $-1.066$. The small changes occur because reasoning levels induce different turnover. Full confidence intervals and portfolio paths are provided in evaluation files that machines can read.

\section{Absolute Performance and Sensitivity to Portfolio Width}

Table~\ref{tab:absolute-performance} supplies economic context for the paired effects. Each cell reports baseline/low reasoning; Gemini's baseline is minimal. Net is mean return after trading costs, the Sharpe ratio scales mean return by return volatility, maximum drawdown (MDD) is the largest loss from a peak to a trough, and turnover is the average fraction of capital traded in one direction. Absolute levels are descriptive historical backtests and do not establish prospective profitability.

\begin{center}
\begin{minipage}{\columnwidth}
\centering
\captionof{table}{Absolute portfolio context at a one way cost of 10 bps. Each pair is baseline/low reasoning; metrics are defined immediately above.}
\label{tab:absolute-performance}
\scriptsize
\resizebox{\columnwidth}{!}{%
\begin{tabular}{llrrrr}
\toprule
Backbone & Information & Net & Sharpe & MDD & Turnover\\
\midrule
DeepSeek & numerical & $-6.376/-0.217$ & $-0.80/-0.14$ & $-32.03\%/-20.52\%$ & $30.83\%/24.48\%$\\
DeepSeek & identifiable news & $-4.386/-4.008$ & $-0.59/-0.52$ & $-27.80\%/-24.46\%$ & $31.19\%/27.77\%$\\
DeepSeek & masked news & $-5.412/-3.963$ & $-0.69/-0.50$ & $-26.68\%/-28.48\%$ & $30.79\%/27.88\%$\\
GPT & numerical & $-3.506/+0.033$ & $-0.46/-0.12$ & $-27.44\%/-23.58\%$ & $26.22\%/26.02\%$\\
GPT & identifiable news & $-1.487/-2.113$ & $-0.27/-0.33$ & $-24.94\%/-29.53\%$ & $27.28\%/26.86\%$\\
GPT & masked news & $-3.483/-3.442$ & $-0.47/-0.46$ & $-25.96\%/-26.36\%$ & $27.61\%/26.88\%$\\
Gemini & numerical & $-4.962/-4.204$ & $-0.57/-0.50$ & $-32.17\%/-34.38\%$ & $26.56\%/26.65\%$\\
Gemini & identifiable news & $+1.417/+1.275$ & $+0.04/+0.03$ & $-24.84\%/-23.46\%$ & $27.93\%/29.59\%$\\
Gemini & masked news & $+2.122/+0.999$ & $+0.08/-0.00$ & $-30.15\%/-21.77\%$ & $28.38\%/29.51\%$\\
\bottomrule
\end{tabular}}
\end{minipage}
\end{center}

Table~\ref{tab:tail-sensitivity} reuses the frozen scores while varying only the number of selected names per side. The result for 10 long and 10 short names remains primary. One of 36 exploratory cells meets the unadjusted rule for material benefit: DeepSeek masked news at 20 names per side. Its HAC 95\% CI is $[0.597,10.942]$ bps/day and its interval from a moving-block bootstrap with five sessions per block is $[0.690,11.223]$. Across the family of 36 cells, however, no HAC or moving-block $p$-value survives Holm adjustment; both adjusted $p$-values for this cell equal 1.000. The sweep therefore demonstrates sensitivity to portfolio width rather than a confirmatory reasoning benefit.

\begin{center}
\begin{minipage}{\columnwidth}
\centering
\captionof{table}{Sensitivity to portfolio width: net return for low relative to baseline reasoning (bps/day) at a one way cost of 10 bps. Columns give names selected per side. An asterisk marks the sole unadjusted cell with material benefit.}
\label{tab:tail-sensitivity}
\scriptsize
\resizebox{\columnwidth}{!}{%
\begin{tabular}{llrrrr}
\toprule
Backbone & Information & 5 & 10 (primary) & 15 & 20\\
\midrule
DeepSeek & numerical & $+5.489$ & $+6.159$ & $+9.239$ & $+5.315$\\
DeepSeek & identifiable news & $-5.362$ & $+0.378$ & $+2.399$ & $+3.909$\\
DeepSeek & masked news & $+1.233$ & $+1.449$ & $+6.206$ & $+5.769^{*}$\\
GPT & numerical & $-6.748$ & $+3.540$ & $+2.891$ & $+0.409$\\
GPT & identifiable news & $-4.980$ & $-0.626$ & $-3.883$ & $-0.708$\\
GPT & masked news & $-2.507$ & $+0.040$ & $+0.631$ & $+2.887$\\
Gemini & numerical & $+12.474$ & $+0.758$ & $-0.468$ & $+0.118$\\
Gemini & identifiable news & $-4.739$ & $-0.142$ & $+1.386$ & $+1.617$\\
Gemini & masked news & $+0.315$ & $-1.123$ & $+0.016$ & $-0.756$\\
\bottomrule
\end{tabular}}
\end{minipage}
\end{center}

\section{Precision and Detectable Effects}

Table~\ref{tab:detectable-effects} reports a design diagnostic based on the observed paired HAC standard error. The \emph{minimum detectable effect} (MDE) is the smallest true effect that a design of this precision would detect with a chosen probability; the two-sided 80\% MDE is $(z_{.975}+z_{.80})\widehat{\mathrm{SE}}$. This calculation summarizes the precision of the realized design under its estimated dependence structure; it is not an equivalence test and does not convert an inconclusive contrast into a zero effect.

\begin{center}
\begin{minipage}{\columnwidth}
\centering
\captionof{table}{HAC precision diagnostics for the nine primary low versus baseline contrasts, in bps/day; MDE is minimum detectable effect.}
\label{tab:detectable-effects}
\small
\resizebox{\columnwidth}{!}{%
\begin{tabular}{llrrr}
\toprule
Backbone & Information & HAC SE & 95\% radius & 80\% MDE\\
\midrule
DeepSeek & numerical & 6.838 & 13.402 & 19.156\\
DeepSeek & identifiable news & 4.802 & 9.411 & 13.452\\
DeepSeek & masked news & 4.192 & 8.217 & 11.745\\
GPT & numerical & 3.097 & 6.071 & 8.678\\
GPT & identifiable news & 2.693 & 5.279 & 7.545\\
GPT & masked news & 2.213 & 4.337 & 6.200\\
Gemini & numerical & 3.567 & 6.991 & 9.993\\
Gemini & identifiable news & 4.074 & 7.986 & 11.415\\
Gemini & masked news & 3.722 & 7.295 & 10.428\\
\bottomrule
\end{tabular}}
\end{minipage}
\end{center}

\section{Agreement Across Repeated Generations}

\emph{Spearman rank correlation} ($\rho$) measures how similarly two generations order all assets. \emph{Tail Jaccard} is the fraction of names shared by their combined selected long and short sets. \emph{Identical tails} records whether both selected sets match exactly.

\begin{center}
\begin{minipage}{\columnwidth}
\centering
\captionof{table}{Mean pairwise generation agreement across the three generation pairs.}
\label{tab:stochastic-agreement}
\small
\resizebox{\columnwidth}{!}{%
\begin{tabular}{lllrrr}
\toprule
Backbone & Information & Level & Score $\rho$ & Tail Jaccard & Identical tails\\
\midrule
GPT & numerical & none & .946 & .642 & .000\\
GPT & numerical & low & .946 & .657 & .000\\
GPT & identifiable news & none & .899 & .591 & .000\\
GPT & identifiable news & low & .911 & .589 & .000\\
GPT & masked news & none & .905 & .567 & .000\\
GPT & masked news & low & .915 & .583 & .000\\
Gemini & numerical & minimal & .858 & .515 & .000\\
Gemini & numerical & low & .741 & .359 & .000\\
Gemini & identifiable news & minimal & .811 & .479 & .000\\
Gemini & identifiable news & low & .692 & .363 & .000\\
Gemini & masked news & minimal & .816 & .461 & .000\\
Gemini & masked news & low & .669 & .355 & .000\\
\bottomrule
\end{tabular}}
\end{minipage}
\end{center}

Each pairwise agreement calculation contains 4,800 asset rows per level and information condition. No audited GPT or Gemini formation date reproduces the exact same tails in any generation pair.

\section{Analysis by Calendar Period}

Table~\ref{tab:calendar-subperiods} partitions the nine primary daily effects of low relative to baseline using predefined calendar boundaries. Values are descriptive subgroup means in bps/day; the annual estimates and inference remain primary.

\begin{center}
\begin{minipage}{\columnwidth}
\centering
\captionof{table}{Effects by calendar period in 2024, in bps/day. H1 is January--June, H2 is July--December, and Q1--Q4 are calendar quarters, assigned by portfolio return date.}
\label{tab:calendar-subperiods}
\scriptsize
\resizebox{\columnwidth}{!}{%
\begin{tabular}{lrrrrrr}
\toprule
Contrast & H1 & H2 & Q1 & Q2 & Q3 & Q4\\
\midrule
DeepSeek numerical & $-10.027$ & $+20.687$ & $-14.004$ & $-6.807$ & $+21.440$ & $+19.922$\\
DeepSeek identifiable news & $+1.290$ & $-0.440$ & $-8.325$ & $+9.073$ & $-3.715$ & $+2.886$\\
DeepSeek masked news & $-3.121$ & $+5.550$ & $+1.628$ & $-6.965$ & $+8.364$ & $+2.693$\\
GPT numerical & $+4.138$ & $+3.002$ & $+10.946$ & $-1.373$ & $+6.866$ & $-0.922$\\
GPT identifiable news & $+2.791$ & $-3.692$ & $+2.837$ & $+2.753$ & $-9.938$ & $+2.653$\\
GPT masked news & $+1.581$ & $-1.342$ & $+2.623$ & $+0.737$ & $+0.157$ & $-2.866$\\
Gemini numerical & $+2.935$ & $-1.197$ & $+6.603$ & $-0.034$ & $-2.617$ & $+0.246$\\
Gemini identifiable news & $-6.085$ & $+5.192$ & $-7.416$ & $-5.008$ & $+7.228$ & $+3.125$\\
Gemini masked news & $+0.776$ & $-2.827$ & $-3.361$ & $+4.125$ & $+12.083$ & $-17.974$\\
\bottomrule
\end{tabular}}
\end{minipage}
\end{center}

DeepSeek numerical has a raw heterogeneity $p=.014$ for H2 minus H1, and Gemini masked news has a raw $p=.045$ for equality across quarters; their values after Holm adjustment across the family of 18 tests are $.246$ and $.757$. No heterogeneity test survives the family correction. The largest quarter accounts for 34.2\%--64.5\% of each contrast's sum of absolute quarterly contributions, so no annual estimate is wholly produced by one quarter. However, only GPT numerical retains the same sign in both halves, and seven annual directions flip when the estimate is recomputed after omitting at least one quarter. These diagnostics show temporal breadth and instability within 2024, not independent evidence across market years.

\section{Protocol Integrity and Claim Boundaries}

Table~\ref{tab:protocol-status} summarizes the fixed study status. All three backbones and low versus baseline contrasts were predetermined, and each backbone evaluation froze its inputs, prompts, schema, failure policy, portfolio mapping, and estimand before outcome evaluation.

\begin{center}
\begin{minipage}{\columnwidth}
\centering
\captionof{table}{Protocol status of the three completed backbone evaluations. Operational amendments did not change reasoning levels or return analysis.}
\label{tab:protocol-status}
\scriptsize
\resizebox{\columnwidth}{!}{%
\begin{tabular}{llll}
\toprule
Backbone & Reasoning contrast & Formation dates & Operational note\\
\midrule
DeepSeek-Flash & none to low & 241 & no amendment that changed treatment\\
GPT-5.6 Luna & none to low & 241 & delivery amendment only; outcomes unseen\\
Gemini 3.1 Flash-Lite & minimal to low & 241 & transport/concurrency only; outcomes unseen\\
\bottomrule
\end{tabular}}
\end{minipage}
\end{center}

Additional disclosures are required for interpretation:
\begin{itemize}
  \item GPT's primary evaluation used 3,120 Batch tasks and 736 synchronous Responses tasks under a frozen delivery amendment made without viewing outcomes. \emph{Delivery mode} means how requests were sent to the provider; because it changes with calendar time, it is not a causal treatment.
  \item Every backbone has three generations for every information condition on the same audit sample of 48 formation dates. The repeated calls measure output and portfolio stability without changing the primary annual estimands or the number of return dates.
  \item Gemini 3.1 Flash-Lite passed the frozen admission gate for identifier fidelity, a check that it reproduces each opaque asset code exactly, before backbone evaluation.
  \item Gemini amendments to state labels, response files, concurrency (the number of simultaneous requests), and orchestration (request scheduling) altered transport only. Prompts, tasks, reasoning levels, failure handling, budgets, and the evaluator were unchanged.
\end{itemize}

\noindent\textbf{Artifact availability and responsible use.}
The authors retain complete evaluation records and plan to release the evaluation code, prompts, result summaries, evaluator hashes, and artifact manifest in a public GitHub repository. Licensed market and news data, together with provider outputs restricted by applicable terms, will not be redistributed. This retrospective simulation placed no live trades and recruited no human participants. It is not investment advice or evidence of prospective profitability. Any public release will preserve the historical backtest label, documented claim boundaries, and applicable licensing restrictions.

\end{document}